\documentclass[12pt,letterpaper]{article}	
\UseRawInputEncoding
\usepackage{expex,marvosym,tabularray,xcolor}

\usepackage{times} 
\usepackage{tipa}
\usepackage{graphicx}
\usepackage{multirow}
\usepackage{subcaption}
\usepackage{natbib}
 	\setcitestyle{semicolon,aysep={},yysep={,},notesep={:}}

\usepackage{lipsum} 
\usepackage{ragged2e}
\RaggedRight

\usepackage[bottom]{footmisc} 
\usepackage{scrextend} 
\deffootnote[.5em]{0em}{1em}{\textsuperscript{\thefootnotemark}\,}

\newcounter{savefootnote}   
\newcounter{symfootnote}
\newcommand{\symfootnote}[1]{%
   \setcounter{savefootnote}{\value{footnote}}%
   \setcounter{footnote}{\value{symfootnote}}%
   \ifnum\value{footnote}>8\setcounter{footnote}{0}\fi%
   \let\oldthefootnote=\thefootnote%
   \renewcommand{\thefootnote}{\fnsymbol{footnote}}%
   \footnote{#1}%
   \let\thefootnote=\oldthefootnote%
   \setcounter{symfootnote}{\value{footnote}}%
   \setcounter{footnote}{\value{savefootnote}}%
}

\usepackage[labelsep=period]{caption}

\usepackage[margin=1.0in]{geometry}
\usepackage[compact]{titlesec}
	\titleformat{\section}[runin]{\normalfont\bfseries}{\thesection.}{.5em}{}[.]
	\titleformat{\subsection}[runin]{\normalfont\scshape}{\thesubsection.}{.5em}{}[.]
	\titleformat{\subsubsection}[runin]{\normalfont\scshape}{\thesubsubsection.}{.5em}{}[.]
\usepackage[colorlinks,allcolors={black},urlcolor={blue}]{hyperref} 

\usepackage[utf8]{inputenc}

\renewenvironment{abstract}{%
\noindent\begin{minipage}{1\textwidth}
\setlength{\leftskip}{0.4in}
\setlength{\rightskip}{0.4in}
\textbf{Abstract.}}
{\end{minipage}}

\newenvironment{keywords}{%
\vspace{.5em}
\noindent\begin{minipage}{1\textwidth}
\setlength{\leftskip}{0.4in}
\setlength{\rightskip}{0.4in}
\textbf{Keywords.}}
{\end{minipage}}

\begin{document} 


\begin{center}
\normalfont\bfseries
Stronger Together: Unleashing the Social Impact of Hate Speech Research
\vskip .5em
\normalfont
    {Sidney Wong\symfootnote{I want to acknowledge Fulbright New Zealand $|$ Te T\={u}\={a}papa M\={a}tauranga o Aotearoa me Amerika and their partnership with the Ministry of Business, Innovation, and Employment $|$ H\={i}kina Whakatutuki for their support through the Fulbright New Zealand Science and Innovation Graduate Award. I want to thank Jonathan Dunn at the University of Illinois Urbana-Champaign for hosting me as a visiting student researcher to undertake this research. Additionally, I want to thank Simon Todd and the Computational Psycholinguistics of Listening and Speaking (CPLS) Lab at the University of California, Santa Barbara, and Abby Walker and the VirginiaTech Language Sciences (VTLx) research group for hosting me and providing feedback on this research. Lastly, I thank the reviewers, organisers, and attendees at their Linguistic Society of America 2025 Annual Conference in Philadelphia for their valuable feedback. Author: Sidney Wong, Geospatial Research Institute; University of Canterbury (\href{mailto:sidney.wong@pg.canterbury.ac.nz}{sidney.wong@pg.canterbury.ac.nz})}}
\vskip .5em
\end{center}

\begin{abstract}
    The advent of the internet has been both a blessing and a curse for once marginalised communities. When used well, the internet can be used to connect and establish communities crossing different intersections; however, it can also be used as a tool to alienate people and communities as well as perpetuate hate, misinformation, and disinformation especially on social media platforms. We propose steering hate speech research and researchers away from pre-existing computational solutions and consider social methods to inform social solutions to address this social problem. In a similar way linguistics research can inform language planning policy, linguists should apply what we know about language and society to mitigate some of the emergent risks and dangers of anti-social behaviour in digital spaces. We argue linguists and NLP researchers can play a principle role in unleashing the social impact potential of linguistics research working alongside communities, advocates, activists, and policymakers to enable equitable digital inclusion and to close the digital divide.
\end{abstract}


\section*{Content Warning} \textit{The following paper makes references to hate speech, offensive language, and violence. All attempts have been made to obfuscate examples of slurs; however, there may still remain traces of unobfuscated examples of slurs in the text.}

\section{Introduction}

    In the last three decades, we have seen an exponential growth into hate speech research with rapid developments in the last decade alone as a result of methodological advancement in computational linguistics and NLP \citep{tontodimamma_thirty_2021}. These advancements have been purported as a valuable resource in policing anti-social behaviour online \citep{rawat_hate_2024}. However, community-minded researchers are beginning to question the benefits of computational solutions in combating hate speech \citep{parker_is_2023}. In order to illustrate some of challenges in the application of hate speech detection systems, we take social media posts from New Zealand as a case study to determine the pitfalls of existing systems. Therefore, our primary research question: can we monitor the increase of hate speech on social media using automatic hate speech systems? In addition to our primary research question, our meta-research question is: if we cannot use automatic hate speech detection systems to monitor hate speech on social media, what are the alternative approaches?

\subsection{Hate Speech Detection}

    While the linguistic and discursive features of hate speech remain poorly defined \citep{guillen-nieto_hate_2023}, the development of automatic hate speech detection systems in NLP has been continuously justified by the purported social benefits of these systems \citep{hovy_social_2016}. Some of the earliest hate speech detection systems using social media language data took an unsupervised learning approach on lexical and syntactic feature representations \citep{chen_detecting_2012}. Other approaches to automatic hate speech detection included semi-supervised topic modelling \citep{xiang_detecting_2012} or supervised modelling \citep{dinakar_common_2012}. 
    
    With the popularisation of language embedding models, automatic hate speech detection is now treated as a supervised text classification task following a standardised pipeline \citep{jahan_systematic_2023}. State-of-the-art hate speech detection models are normally developed with the following pipeline: a) \textit{data set collection and preparation}, which involves collecting either real-world or synthetic instances of hate speech; b) \textit{feature engineering}, which involves manipulating and transforming instances of hate speech in preparation for language modelling; c) \textit{model training}, which involves developing a hate speech detection system with machine learning algorithms; and lastly, c) \textit{model evaluation}, which involves producing model performance metrics to determine the statistical validity of the system.

    In light of these technological advances, there is little evidence that these efforts are being deployed to support target communities. This is because these hate speech detection systems are seldom used by non-profit organisations (NGOs) or policy-makers in combatting anti-social behaviour in digital spaces \citep{parker_is_2023}. In an attempt to develop and to deploy one of these systems to support municipal elections in Finland, \citet{laaksonen_datafication_2020} found that these systems became an unnecessary distraction for researchers in collaborating with their stakeholders due to the `datafication' of hate speech. This conclusion reinforces the finding in \citet{parker_is_2023} that hate speech detection as a methodological solution may not be meeting the needs of target communities.
    
    A systematic review of 48 open-source hate speech detection datasets found that these systems provided more benefit to NLP researchers than target communities \citep{wong_what_2024}. The review was conducted in line with the \textit{Responsible Natural Language Processing} \citep{behera_responsible_2023} conceptual model which outlines eight principles for ethical and responsible research in NLP. While the hate speech detection systems reviewed all scored highly in areas such as reliability, interrogation, and in privacy and security through research activities like shared tasks, the results of the review suggested that these systems failed to meet their ethical obligations in the principles of accountability and fairness due to their lack of engagement with target communities.

\subsection{Bias and Limitations}
        
    \citet{arango_hate_2022} argued that pre-existing approaches to model evaluation in NLP over-estimate the performance of current state-of-the-art automatic hate speech detection systems which are especially susceptible to bias. \citet{davidson_racial_2019} tested for racial bias in five hate speech detection systems for Twitter using (including \citealp{waseem_are_2016}; \citealp{waseem_are_2016}; \citealp{davidson_automated_2017}; \citealp{golbeck_large_2017}; and \citealp{founta_large_2018}) by using \citet{blodgett_demographic_2016} data as a proxy for race. The authors found consistent, systematic, and substantial racial bias across all systems which disproportionately negatively African-American English due to the sampling procedures used to collect training data. 
    
    The results from \citet{davidson_automated_2017} suggest that relying on lexical features and researcher judgements may ignore other social or linguistic processes such as reclamation or reappropriation of slur-like lexical feature. Furthermore, racial bias can be introduced throughout the development pipeline such as during the data annotation process \citep{sap_risk_2019}. Additionally, these systems were found to lack cultural awareness \citep{lee_hate_2023} which means we cannot simply apply hate speech detection systems in one language condition to another language condition due to culture-specific biases (\citealp{zhou_cross-cultural_2023}; \citealp{wong_sociocultural_2024})
    
\subsection{Summary}

    The lack of engagement from NGOs and policy-makers in hate speech detection research, and the methodological limitations in developing these systems highlight the difficulties in developing large and varied automatic hate speech detection systems that are theoretically-informed while minimising bias \citep{vidgen_directions_2020}. \cite{guillen-nieto_hate_2023} argued that linguists may play an essential role in combatting hate speech with our intuitive knowledge of language. More so, \citet{alonso_alemany_bias_2023} argued that bias analysis should not rely on computational approaches which inadvertently oversimplify and reduce these complex social processes. Therefore, the current paper offers linguists an opportunity to consider how we can collectively unleash the social impact of hate speech research with target communities.

\section{Case Study: New Zealand}

    New Zealand is a predominately English-speaking island nation in the South Pacific Ocean. Recent events have increased the public awareness on the need to regulate and limit hate speech in the digital spaces of New Zealand \citep{hoverd_christchurch_2020}.
    
    On Friday, 15 March 2019, an Australian national carried out two consecutive mass shootings at two mosques in Christchurch killing 51 people. Motivated by white supremacist and alt-right extremist ideologies, this incident increased the public awareness between linking the rise of hate, harassment, and terrorism and unregulated social media platforms \citep{hoverd_christchurch_2020}. As part of the Royal Commission of Inquiry into the terrorist attack, the findings provided the New Zealand public its first definition of hate speech to mean ``speech that expresses hostility towards, or contempt for, people who share a characteristic'' \citep{royal_commission_of_inquiry__into_the_terrorist_attack_on_christchurch_masjidain_on_15_march_2019_ko_2020}. One recommendation from the Inquiry was to incorporate hate crime and hate speech into New Zealand's existing legislative framework in order to promote social cohesion and to close the digital divide for vulnerable target communities.

    It was during this time, the Disinformation Project (2020-2024) - a New Zealand-based independent research group providing best practice monitoring, research, and consulting on misinformation and its impacts - published a report finding that hate speech on social media directed towards LGBTQ+ communities has increased in both volume and tone \citep{hattotuwa_transgressive_2023}. These insights were largely based on qualitative accounts across different social media platforms. The domestic increase of hate speech has been attributed to the corroboration of disinformation and misinformation spread by alternative media (alt-media) platforms including climate change deniers, anti-COVID vaccination mandate activists, and proponents of the Gender-critical feminist movement \citep{clark_intersections_2023}. 
    
    While much of the hate speech has been confined to digital platforms, such as Telegram, there is growing recognition that hate speech (including misinformation and disinformation) are having real-world impacts on target communities \citep{hoverd_christchurch_2020}. Recent work has shown that linguistic behaviour on social media platforms can be linked to real-world events such as a decrease of linguistic diversity during the Covid-19 Pandemic \citep{dunn_measuring_2020}. Despite the on-going threat of hate speech and hate crimes to social cohesion, the New Zealand Government has shelved proposed legislative reforms on hate speech as of February 2023 citing the need to redirect public funds for economic growth.

\section{Methodology}

    \begin{figure}[t]
        \centering
        \includegraphics[width=\columnwidth]{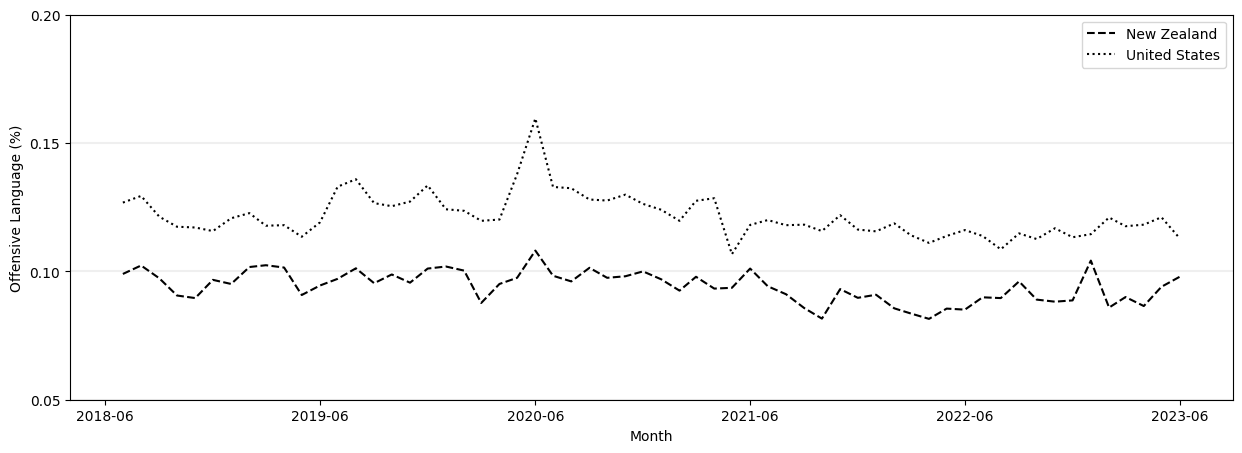}
        \caption{\label{fig:monthly_proportion} Proportion of offensive language over time (monthly).}
    \end{figure}

    With this sociolinguistic and political context of New Zealand in mind, we develop a simple text classification model. While we are primarily interested in the efficacy of automatic hate speech detection at a local-level, the purpose of the country-level analysis is to determine possible geographic bias in either the open-source training data or the pre-trained language models. Based on the recent events within New Zealand, we would expect to see an increase of hate speech and offensive language in the periods following the Mosque attacks centred on Christchurch (15 March 2019), the first nationwide lockdown restrictions due to Covid-19 (25 March 2019), the anti-mandate and anti-lockdown protests on the grounds of parliament in Wellington (6 February 2022), and the tour of gender-critical activist Kellie-Jay Keen-Minshull in Auckland and Wellington (25 March 2023).

\subsection{Hypothesis}

    Our primary research question is: Can we monitor the increase of hate speech on social media using automatic hate speech detection. Therefore, our hypothesis is that the rate of hate speech has increased on social media in New Zealand. This means the null hypothesis is that there has been no change in the rate of hate speech on social media in New Zealand.

\subsection{Data Sources}
     
    We use geo-referenced social media language data for analysis and domain adaptation of the pre-trained language models which we discuss further in Section \ref{subsec:model}. The source of the geo-referenced X (Twitter) data is from the \textit{Corpus of Global Language Use} (CGLU; \citealp{dunn_mapping_2020}). The geo-referenced posts (tweets) were all produced within a 50-kilometre radius from each of the 100 data collection points for all countries in the CGLU. Data collection has been on-going since June 2018. In the country-level analysis, we extract a sample of 10,000 English language posts (tweets) per month from the United States and New Zealand between June 2018 to May 2023. In the local-level analysis, we extract a sample of 1,000 English language posts (tweets) per month from each of the 100 data collection points across New Zealand between June 2018 to May 2023. In addition to the sample posts (tweets), we also extract an additional sample of 50,000 English-language posts (tweets) for domain adaptation.

    The source of the open source hate speech training data is derived from \citet{davidson_automated_2017} who took a keyword approach to collecting samples of hate speech and offensive language on X (Twitter). With Hatebase\footnote{\href{https://hatebase.org/}{hatebase.org}} as the primary source, \citet{davidson_automated_2017} established a candidate list of 178 one to four word \textit{n}-grams. Using the Twitter API, 85.4 million posts (tweets) were extracted from which a sample of approximately 25,000 posts (tweets) were manually coded through CrowdFlower with one of the three categories: hate speech; offensive but not hate speech; or neither offensive nor hate speech. Each post (tweet) was coded by at least three or more people and the CrowdFlower intercoder-agreement score was 92\%. The best performing text classification model in \citet{davidson_automated_2017} yielded a precision of 0.91, recall of 0.90, and $F_1$-score of 0.90. We retrieved the hate speech training data from GitHub repository\footnote{\href{https://github.com/t-davidson/hate-speech-and-offensive-language}{t-davidson/hate-speech-and-offensive-language}}. The final training data set had 1,430 instances of hate speech (5.8\%), 19,190 instances of offensive language (77.4\%), and 4,163 instances in the residual category (16.8\%).

    \begin{figure}
      \centering
      \subfloat[New Zealand\label{fig:linear_model}]{\includegraphics[height=0.35\textwidth]{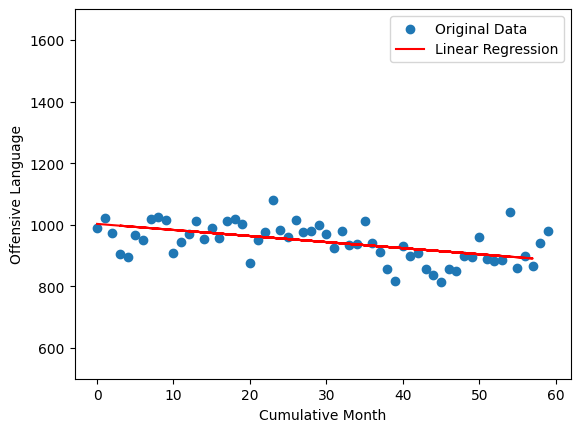}}\quad \subfloat[United States\label{fig:nz}]{\includegraphics[height=0.35\textwidth]{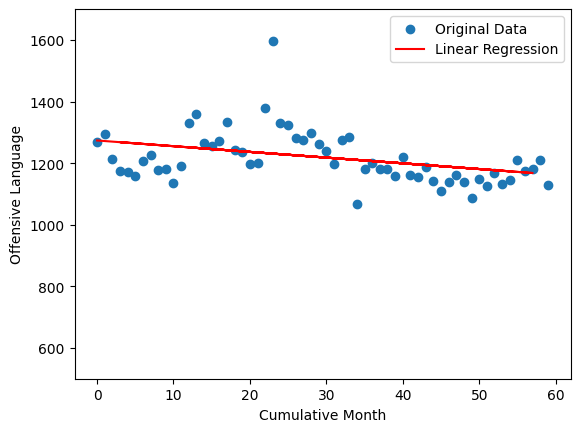}}
      \caption{Linear Regression model results.}
      \label{fig:us}
    \end{figure}

\subsection{Model Development}
\label{subsec:model}

    As part of our analysis, we train two multi-class classification models. For our first baseline text classification model, we use the classification class from the simpletransformers\footnote{\href{https://simpletransformers.ai/}{simpletransformers.ai}} library. We split the training data into train, validation, and test sets for model development. We did not process the training data for class imbalance. We fine-tune XLM-Roberta \cite{liu_roberta_2019} - a multilingual pre-trained large language model - with the labelled training data from \citet{davidson_automated_2017}. We trained the model for eight iterations and evaluated the training for every 500 steps. In our New Zealand-specific model, we pretrained XLM-Roberta \cite{liu_roberta_2019} with 50,000 samples of posts (tweets) from New Zealand. We then trained a multi-class text classification model using the three-class hate speech detection data set produced by \citet{davidson_automated_2017}.

\section{Results}
\label{sec:results}

    We first discuss the results of the country-level analysis by comparing the results between New Zealand and the United States. We then discuss the results of the local-level analysis for the regions and cities of New Zealand.

\subsection{Country-Level Analysis}
\label{subsec:country}

    \begin{figure}[t]
        \centering
        \includegraphics[width=\columnwidth]{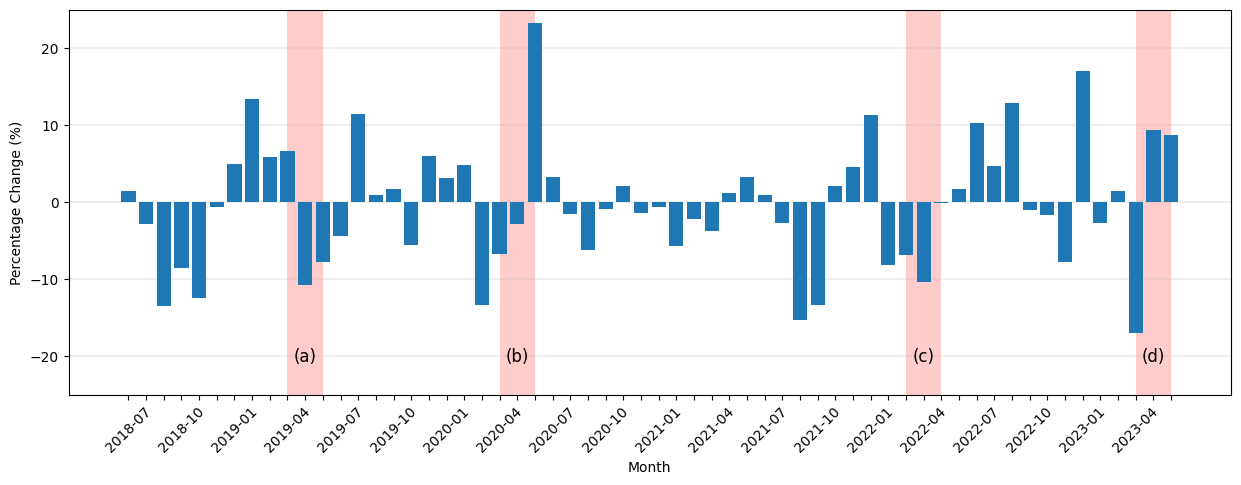}
        \caption{\label{fig:change} Percentage Change (Period = 3) for New Zealand with the time periods of interest highlighted in red where (a) was the period following the Mosque attack, (b) was the period following the introduction of the nationwide lockdown, (c) was the period following the anti-mandate and anti-vaccine protests, and (d) was the period following the speaking tour of anti-transgender activist Kellie-Jay Keen-Minshull.}
    \end{figure}

    The accuracy of our baseline text classification model, which we will refer to as the hate speech detection model going forward, with no domain adaptation on social media language data was 0.89, the macro average $F_1$-score was 0.59, and the weighted average $F_1$-score was 0.86. The macro average $F_1$-score is where all classes equally contribute to the averaged $F_1$-score, whereas the weighted average is where the classes are weighted by size. Based on the macro averaged $F_1$-score, the performance of the hate speech detection model on slightly above chance ($>50\%$) which suggest poor performance on the minority classes (i.e., hate speech).
    
    With reference to the model performance metrics, we applied the hate speech detection model on unseen social media language data for New Zealand and the United States. The model did not identify any instances of hate speech; however, it did identify instances of offensive language. This was expected based on the model performance metrics, which performed poorly on the minority classes. The proportion of offensive language per 10,000 social media posts for New Zealand and the United States are presented in Figure \ref{fig:monthly_proportion}. As shown in Figure \ref{fig:monthly_proportion}, we observed that the proportion of offensive language was much higher in the United States than New Zealand. Both countries observed a maximum peak in June 2020, while New Zealand observed a secondary peak in January 2023.

    When we calculated the Pearson Correlation Coefficient (PCC) between the rate of offensive language between New Zealand and the United States, we found a statistically significant moderate positive correlation of $0.596$ ($p<0.05$) between the two countries. As we are interested in the relationship between the rate of offensive language over time, we converted the date variable to an ordinal variable or the cumulative months from the starting period of June 2018. We then calculated the PCC for the rate of offensive language and the cumulative months for New Zealand which was $-0.502$ ($p<0.05$) and for the United States which was $-0.401$ ($p<0.05$). This means there was a statistically significant moderate negative correlation between offensive language and cumulative months for both country conditions which suggests the rate of offensive language was decreasing over time. We can observe this trend this by visually inspecting Figure \ref{fig:monthly_proportion} which suggests the rate of offensive language is decreasing over time.

    While the PCC measures were sufficient in allowing us to test our hypothesis, we wanted to confirm our findings by running a linear regression model for each country condition. We did this by splitting the data into train and test sets ($75:25$). The results for New Zealand are shown in Figure \ref{fig:nz}. The Mean Squared Error (MSE) was 3,289.50, the $R$-squared was 0.855, the coefficient was -1.970, and the intercept was 1,002.713. The model performance metrics suggest a poor fit. In order to retrieve the $p$-value, we added a constant to the independent variable to include an intercept. While the model suggested a poor fit, we can confirm the statistically significant ($p<0.05$) moderate negative relationship between the rate of offensive language over time. The results for the United States offered a similar interpretation with an MSE of 3,824.271, an $R$-squared of 0.276, a coefficient of -1.855, and an intercept of 1,273.54.

    Finally, we wanted to impressionistically determine if there was a relationship between the rate of hate speech (in this case, offensive language only) and recent events in New Zealand. We calculated the percentage change based on the last quarter for New Zealand in Figure \ref{fig:change} with the time periods of interest highlighted in red. We did not observe an increase of offensive language in the period following the Mosque attack (a) or in the period following the anti-mandate and anti-vaccine protests (c); however, we can observe an increase of offensive language in the period following the introduction of the nationwide lockdown (b) and the period following the speaking tour of anti-transgender activist Kellie-Jay Keen-Minshull (d).

    \begin{table}
        \centering
        \begin{tabular}{lcc}
            \hline
            Period & Hate Speech & Offensive Language\\
            \hline
            June 2018 - May 2019 & 540 & 29,373 \\
            June 2019 - May 2020 & 581 & 33,279 \\
            June 2020 - May 2021 & 577 & 34,398 \\
            June 2021 - May 2022 & 588 & 30,563 \\
            June 2022 - May 2023 & 486 & 27,569 \\
            \hline
         \end{tabular}
         \caption{\label{tab:nz_compare} Instances of hate speech and offensive language observed in New Zealand grouped into 12-month periods from 1 June to 31 May.}
     \end{table}

    \begin{figure*}[t]
        \centering
        \includegraphics[width=\columnwidth]{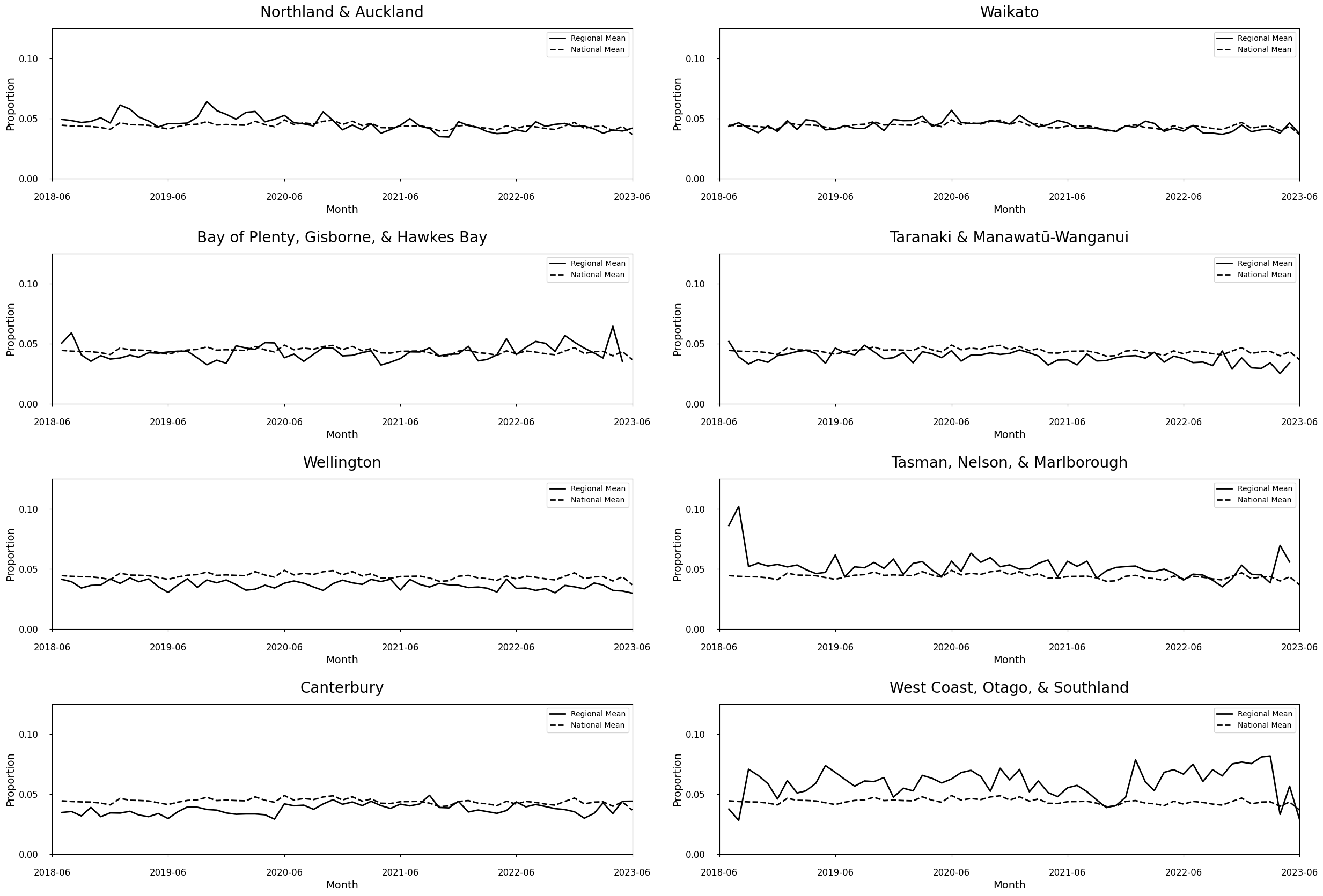}
        \caption{\label{fig:proportion_local} Proportion of hate speech and offensive language by regions.}
    \end{figure*}

\subsection{Local-Level Analysis}

    We now turn our focus to our local-level analysis for New Zealand only. The accuracy of the hate speech detection model with domain adaptation using social media language data was 0.90, the macro average $F_1$-score was 0.77, and the weighted average $F_1$-score was 0.90. These results suggest that domain adaptation improved the performance of the hate speech detection model when compared with our baseline model where we did not include domain adaptation. The following analysis is largely impressionistic.
    
    Once we had the model performance metrics, we applied the adapted hate speech detection model on unseen social media data on city-level samples of New Zealand. In contrast to the baseline model, the model did detect instances of both hate speech and offensive language. Due to the low frequency detected hate speech and offensive language, we grouped the instances of hate speech and offensive language for the whole country in Table \ref{tab:nz_compare}. Instances of hate speech remained constant for the six periods with the rate of between 486 and 588 instances of hate speech and between 27,569 and 34,398 instances of offensive language. The greatest occurrence of hate speech was in the period between June 2021 to May 2022 while the greatest occurrence of offensive language was in the following period between June 2022 to May 2023.

    After determining the level of granularity in our analysis appropriate to the instances of predicted outputs, we grouped the instances of hate speech and offensive languages into geographic regions over time. These can be found in Figure \ref{fig:proportion_local} where we have plotted the combined proportion of hate speech and offensive language. To aid interpretation, we have included the country-level baseline as a guide. Based on the plot, we can observe that urban regions (such as Wellington and Christchurch) were consistently below the national mean while rural regions (such as Bay of Plenty, Gisborne, and Hawke's Bay; Tasman, Nelson, and Marlborough; and the West Coast, Otago, and Southland) were consistently above the national mean. The stable proportions in the predicted outputs does call into question if some other linguistic processes are at play.
        
\begin{figure}
      \centering
      \subfloat[Hate Speech\label{fig:hate}]{\includegraphics[height=0.25\textwidth]{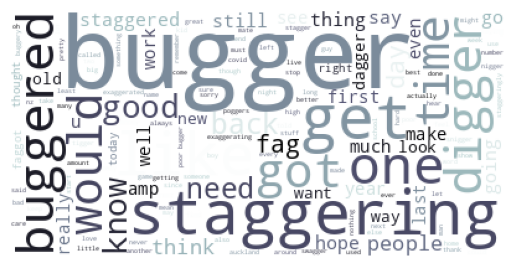}}\quad \subfloat[Offensive Language\label{fig:offensive}]{\includegraphics[height=0.25\textwidth]{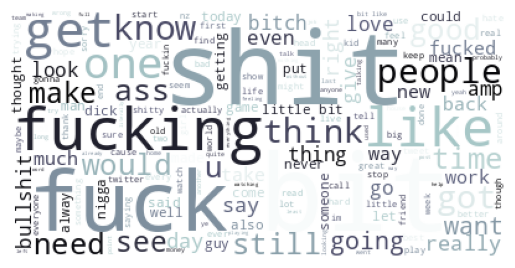}}
      \caption{Predicted instances of hate speech and offensive language for New Zealand.}
      \label{fig:wordcloud}
    \end{figure}

    The trends observed in Figure \ref{fig:proportion_local} were unexpected. Coupled with the stable rate of hate speech and offensive language as shown in Table \ref{tab:nz_compare}, it made us question the validity of our model at the local-level. Therefore, we carried out a topic analysis where we have visualised the predicted outputs of hate speech (\ref{fig:hate}) and offensive language (\ref{fig:offensive}) for New Zealand as presented in Figure \ref{fig:wordcloud}. While our hate speech detection model was able to identify offensive language (as shown on Subfigure \ref{fig:offensive} on the basis on lexical items such as \textit{shit}, \textit{fuck}, and \textit{fucking}, it failed to identify hate speech with reference to Subfigure \ref{fig:hate}. Instead, lexical items that were not present in the training data (such as \textit{bugger}, \textit{staggering}, \textit{buggered}, \textit{digger}) were being (mis)identified as a hate speech.

\section{Discussion and Conclusion}

    We now revisit our primary research question which was: can we monitor the increase of hate speech on social media using automatic hate speech detection. Based on the results for the country-level analysis, we not only reject our null hypothesis that there has been no change in the rate of hate speech on social media in New Zealand, but also our hypothesis that the rate of hate speech has increased on social media in New Zealand. In the contrary, the results suggest that the rate of hate speech on social media in New Zealand (and the United States) has decreased. This result counters the findings from \citet{hattotuwa_transgressive_2023} which found that hate speech has increased in both rate and volume for New Zealand.

    We will first address some of the limitations of our data and methodology before addressing the downstream limitations as a result of the existing hate speech detection pipeline. Firstly, we only took a limited sample of social media posts from the CGLU for each country. Secondly, only a small number of users enable georeferencing on Twitter which means we do not have a representative sample of users from both country conditions. Thirdly, we have not developed evaluation data from the samples to determine if the predicted instances of hate speech or offensive language were indeed hate speech or offensive language.

    Had we taken a the results from the country-level analysis for granted then we would not have been able to uncover the issues of our methodology in the local-level analysis especially with the (mis)identified instances of hate speech. In this case, the model has overfit on lexical items which share \textit{n}-gram features with slurs in the training data (such as \textit{bugger}, \textit{staggering}, \textit{buggered}, \textit{digger}). We argue that treating automatic hate speech detection as a text classification problem pressuposses that instances of hate speech readily appear on social media platforms, when in reality, most social platforms employ their own content moderation processes. This means hate speech detection training data developed solely on lexical features (i.e., slurs) are made redundant when the features are no longer present.

    We further argue the poor performance of our hate speech detection system on unseen social media language data is because the training data produced by \citet{davidson_automated_2017} was not designed based on the social, political, or linguistic context of New Zealand. We would like to stress that \citet{davidson_automated_2017} in our misapplication of the training data. We agree with \citet{alonso_alemany_bias_2023} that model performance metrics, like $F_1$-scores, are less useful in determining bias in hate speech detection systems as bias is inherently a social and linguistic problem. It also brings into question the usefulness of the existing model development pipeline that does not account for evaluating the validity of the model once it has been deployed in unseen samples of social media language data.

    Though we can develop a hate speech detection model on open-source training data, we question the usefulness in the outputs of these models. In lieu of developing a hate speech data set that is specific to the New Zealand context (including the drawbacks of developing more hate speech detection training data that may cause more harm than benefits for target communities), we now refer back to our meta-research question which was: if we cannot use automatic hate speech detection systems to monitor hate speech on social media, what are the alternative approaches? We will discuss these alternatives in the next section.

\subsection{Alternative Approaches}

    Some NLP researchers are moving away from classification-based models towards \textit{counter speech} generation to combat hate speech. Once again, these require researcher judgements and do not derive from the community of interest. In this section, we discuss emergent discursive strategies in hate speech discourse where linguists can provide the most support in monitoring and detecting hate speech. With reference to the organised panel \textit{Language, conflict, and peace-making: contributions from the linguistics and the philosophy of language} at the 2025 LSA Annual Meeting \citep{tirrell_language_2025}, hate speech is only one form of languages of conflict.

    One form of internet language is \textit{Algospeak}, which is a portmanteau of \textit{Algorithm} and \textit{Speak}. This type of language refers to lexical features used to evade automated moderation algorithms \cite{steen_you_2023}. Online users may use forms of Algospeak to discuss sensitive topics. This form of language can be directly linked to language variation and change. Some examples of Algospeak include \textit{(to) unalive} (`to kill; kill; dead; or suicide'); \textit{seggs} (`sex; sexual intercourse'); \textit{yt}  (`white people'); and \textit{Opposite of Love} (`hate'). The development of Algospeak can be traced based on existing models of word formation processes.

    More specific to hate speech are \textit{dog whistles} which refer to coded or suggest (i.e., indirect) language in political messaging to garner support from one group without provoking an opposing group. One example is the bathroom/restroom/toilet discourse used to advance legislation in some jurisdictions which disproportionately impacts trans women. With the support of topic models and critical discourse analysis, linguists may be able to determine the intention of dog whistles to differentiate them from other forms of language-use.

    One final example of internet language is, \textit{Voldermorting}, which much like the fictional character refers to the strategy of not referring to a person by name (i.e., 'He who must not be named' and 'You know who') \cite{van_der_nagel_networks_2018}. One notable example is the censorship of Winnie-the-Pooh in the People's Republic of China where the character was compared to the General Secretary of the Communist Part, Xi Jinping. A more recent example is the Voldermorting of Elon Musk as the anagram, \textit{Leon Skum}.

    These are only some forms of internet language and languages of conflict present in social media language-use. The question remains how linguists and NLP researchers can incorporate these discursive strategies in hate speech detection pipelines. By moving the attention away from hate speech detection, linguists may be able to play a crucial role in combatting not only hate speech, but also digital exclusion.

\setlength{\bibsep}{0pt plus 0.3ex}
\setlength{\bibhang}{0.3in}	
\titleformat{\section}{\normalfont\bfseries}{\thesection}{.5em}{} 

\bibliographystyle{sp-lsa.bst}	
\newcommand{\doi}[1]{\href{https://doi.org/#1}{https://doi.org/#1}} 
\bibliography{references} 

\begin{thebibliography}{35}
\providecommand{\natexlab}[1]{#1}
\providecommand{\url}[1]{\texttt{#1}}
\providecommand{\urlprefix}{}
\expandafter\ifx\csname urlstyle\endcsname\relax
  \providecommand{\doi}[1]{doi:\discretionary{}{}{}#1}\else
  \providecommand{\doi}{doi:\discretionary{}{}{}\begingroup \urlstyle{rm}\Url}\fi

\bibitem[{Alonso~Alemany et~al.(2023)Alonso~Alemany, Benotti, Maina, Gonzalez, Martínez, Busaniche, Halvorsen, Rojo \& Rajngewerc}]{alonso_alemany_bias_2023}
Alonso~Alemany, Laura, Luciana Benotti, Hernán Maina, Lucía Gonzalez, Lautaro Martínez, Beatriz Busaniche, Alexia Halvorsen, Amanda Rojo \& Mariela Rajngewerc. 2023.
\newblock Bias assessment for experts in discrimination, not in computer science.
\newblock In Sunipa Dev, Vinodkumar Prabhakaran, David Adelani, Dirk Hovy \& Luciana Benotti (eds.), \emph{Proceedings of the {First} {Workshop} on {Cross}-{Cultural} {Considerations} in {NLP} ({C3NLP})}, 91--106. Dubrovnik, Croatia: Association for Computational Linguistics.
\newblock \doi{10.18653/v1/2023.c3nlp-1.10}.

\bibitem[{Arango et~al.(2022)Arango, Pérez \& Poblete}]{arango_hate_2022}
Arango, Aymé, Jorge Pérez \& Barbara Poblete. 2022.
\newblock Hate speech detection is not as easy as you may think: {A} closer look at model validation (extended version).
\newblock \emph{Information Systems} 105. 101584.
\newblock \doi{10.1016/j.is.2020.101584}.

\bibitem[{Behera et~al.(2023)Behera, Bala, Rana \& Irani}]{behera_responsible_2023}
Behera, Rajat~Kumar, Pradip~Kumar Bala, Nripendra~P. Rana \& Zahir Irani. 2023.
\newblock Responsible natural language processing: {A} principlist framework for social benefits.
\newblock \emph{Technological Forecasting and Social Change} 188. 122306.
\newblock \doi{10.1016/j.techfore.2022.122306}.

\bibitem[{Blodgett et~al.(2016)Blodgett, Green \& O'Connor}]{blodgett_demographic_2016}
Blodgett, Su~Lin, Lisa Green \& Brendan O'Connor. 2016.
\newblock Demographic {Dialectal} {Variation} in {Social} {Media}: {A} {Case} {Study} of {African}-{American} {English}.
\newblock In Jian Su, Kevin Duh \& Xavier Carreras (eds.), \emph{Proceedings of the 2016 {Conference} on {Empirical} {Methods} in {Natural} {Language} {Processing}}, 1119--1130. Austin, Texas: Association for Computational Linguistics.
\newblock \doi{10.18653/v1/D16-1120}.

\bibitem[{Chen et~al.(2012)Chen, Zhou, Zhu \& Xu}]{chen_detecting_2012}
Chen, Ying, Yilu Zhou, Sencun Zhu \& Heng Xu. 2012.
\newblock Detecting {Offensive} {Language} in {Social} {Media} to {Protect} {Adolescent} {Online} {Safety}.
\newblock In \emph{2012 {International} {Conference} on {Privacy}, {Security}, {Risk} and {Trust} and 2012 {International} {Confernece} on {Social} {Computing}}, 71--80.
\newblock \doi{10.1109/SocialCom-PASSAT.2012.55}.

\bibitem[{Clark \& Stoakes(2023)}]{clark_intersections_2023}
Clark, Byron \& Emanuel Stoakes. 2023.
\newblock Intersections of influence: {Radical} conspiracist 'alt-media' narratives and the climate crisis in {Aotearoa}.
\newblock \emph{Pacific Journalism Review} 29(1/2). 12--26.
\newblock \doi{10.24135/pjr.v29i1and2.1308}.

\bibitem[{Davidson et~al.(2019)Davidson, Bhattacharya \& Weber}]{davidson_racial_2019}
Davidson, Thomas, Debasmita Bhattacharya \& Ingmar Weber. 2019.
\newblock Racial {Bias} in {Hate} {Speech} and {Abusive} {Language} {Detection} {Datasets}.
\newblock In Sarah~T. Roberts, Joel Tetreault, Vinodkumar Prabhakaran \& Zeerak Waseem (eds.), \emph{Proceedings of the {Third} {Workshop} on {Abusive} {Language} {Online}}, 25--35. Florence, Italy: Association for Computational Linguistics.
\newblock \doi{10.18653/v1/W19-3504}.

\bibitem[{Davidson et~al.(2017)Davidson, Warmsley, Macy \& Weber}]{davidson_automated_2017}
Davidson, Thomas, Dana Warmsley, Michael Macy \& Ingmar Weber. 2017.
\newblock Automated {Hate} {Speech} {Detection} and the {Problem} of {Offensive} {Language}.
\newblock In \emph{Proceedings of the {International} {AAAI} {Conference} on {Web} and {Social} {Media}}, vol.~11, 512--515.
\newblock \doi{10.1609/icwsm.v11i1.14955}.

\bibitem[{Dinakar et~al.(2012)Dinakar, Jones, Havasi, Lieberman \& Picard}]{dinakar_common_2012}
Dinakar, Karthik, Birago Jones, Catherine Havasi, Henry Lieberman \& Rosalind Picard. 2012.
\newblock Common {Sense} {Reasoning} for {Detection}, {Prevention}, and {Mitigation} of {Cyberbullying}.
\newblock \emph{ACM Trans. Interact. Intell. Syst.} 2(3). 18:1--18:30.
\newblock \doi{10.1145/2362394.2362400}.

\bibitem[{Dunn(2020)}]{dunn_mapping_2020}
Dunn, Jonathan. 2020.
\newblock Mapping languages: the {Corpus} of {Global} {Language} {Use}.
\newblock \emph{Language Resources and Evaluation} 54(4). 999--1018.
\newblock \doi{10.1007/s10579-020-09489-2}.

\bibitem[{Dunn et~al.(2020)Dunn, Coupe \& Adams}]{dunn_measuring_2020}
Dunn, Jonathan, Tom Coupe \& Benjamin Adams. 2020.
\newblock Measuring {Linguistic} {Diversity} {During} {COVID}-19.
\newblock In \emph{Proceedings of {The} {Fourth} {Workshop} on the {Fourth} {Workshop} on {Natural} {Language} {Processing} and {Computational} {Social} {Science}}, \doi{10.18653/v1/P17}.

\bibitem[{Founta et~al.(2018)Founta, Djouvas, Chatzakou, Leontiadis, Blackburn, Stringhini, Vakali, Sirivianos \& Kourtellis}]{founta_large_2018}
Founta, Antigoni, Constantinos Djouvas, Despoina Chatzakou, Ilias Leontiadis, Jeremy Blackburn, Gianluca Stringhini, Athena Vakali, Michael Sirivianos \& Nicolas Kourtellis. 2018.
\newblock Large {Scale} {Crowdsourcing} and {Characterization} of {Twitter} {Abusive} {Behavior}.
\newblock In \emph{Proceedings of {theTwelfth} {International} {AAAI} {Conference} on {Web} and {Social} {Media}}, vol.~12, Palo Alto, CA: Public Knowledge Project.
\newblock \doi{10.1609/icwsm.v12i1.14991}.

\bibitem[{Golbeck et~al.(2017)Golbeck, Ashktorab, Banjo, Berlinger, Bhagwan, Buntain, Cheakalos, Geller, Gergory, Gnanasekaran, Gunasekaran, Hoffman, Hottle, Jienjitlert, Khare, Lau, Martindale, Naik, Nixon, Ramachandran, Rogers, Rogers, Sarin, Shahane, Thanki, Vengataraman, Wan \& Wu}]{golbeck_large_2017}
Golbeck, Jennifer, Zahra Ashktorab, Rashad~O. Banjo, Alexandra Berlinger, Siddharth Bhagwan, Cody Buntain, Paul Cheakalos, Alicia~A. Geller, Quint Gergory, Rajesh~Kumar Gnanasekaran, Raja~Rajan Gunasekaran, Kelly~M. Hoffman, Jenny Hottle, Vichita Jienjitlert, Shivika Khare, Ryan Lau, Marianna~J. Martindale, Shalmali Naik, Heather~L. Nixon, Piyush Ramachandran, Kristine~M. Rogers, Lisa Rogers, Meghna~Sardana Sarin, Gaurav Shahane, Jayanee Thanki, Priyanka Vengataraman, Zijian Wan \& Derek~Michael Wu. 2017.
\newblock A {Large} {Labeled} {Corpus} for {Online} {Harassment} {Research}.
\newblock In \emph{Proceedings of the 2017 {ACM} on {Web} {Science} {Conference}}, 229--233. New York, NY: Association for Computing Machinery.
\newblock \doi{10.1145/3091478.3091509}.

\bibitem[{Guillén-Nieto(2023)}]{guillen-nieto_hate_2023}
Guillén-Nieto, Victoria. 2023.
\newblock \emph{Hate {Speech}: {Linguistic} {Perspectives}}.
\newblock De Gruyter Mouton.

\bibitem[{Hattotuwa et~al.(2023)Hattotuwa, Hannah \& Taylor}]{hattotuwa_transgressive_2023}
Hattotuwa, Sanjana, Kate Hannah \& Kayli Taylor. 2023.
\newblock Transgressive transitions: {Transphobia}, community building, bridging, and bonding within {Aotearoa} {New} {Zealand}’s disinformation ecologies march-{April} 2023.
\newblock Tech. rep. The Disinformation Project New Zealand.

\bibitem[{Hoverd et~al.(2020)Hoverd, Salter \& Veale}]{hoverd_christchurch_2020}
Hoverd, William~James, Leon Salter \& Kevin Veale. 2020.
\newblock The {Christchurch} {Call}: insecurity, democracy and digital media - can it really counter online hate and extremism?
\newblock \emph{SN Social Sciences} 1(1). 2.
\newblock \doi{10.1007/s43545-020-00008-2}.

\bibitem[{Hovy \& Spruit(2016)}]{hovy_social_2016}
Hovy, Dirk \& Shannon~L. Spruit. 2016.
\newblock The {Social} {Impact} of {Natural} {Language} {Processing}.
\newblock In Katrin Erk \& Noah~A. Smith (eds.), \emph{Proceedings of the 54th {Annual} {Meeting} of the {Association} for {Computational} {Linguistics} ({Volume} 2: {Short} {Papers})}, 591--598. Berlin, Germany: Association for Computational Linguistics.
\newblock \doi{10.18653/v1/P16-2096}.

\bibitem[{Jahan \& Oussalah(2023)}]{jahan_systematic_2023}
Jahan, Md~Saroar \& Mourad Oussalah. 2023.
\newblock A systematic review of hate speech automatic detection using natural language processing.
\newblock \emph{Neurocomputing} 546. 126232.
\newblock \doi{10.1016/j.neucom.2023.126232}.

\bibitem[{Laaksonen et~al.(2020)Laaksonen, Haapoja, Kinnunen, Nelimarkka \& Pöyhtäri}]{laaksonen_datafication_2020}
Laaksonen, Salla-Maaria, Jesse Haapoja, Teemu Kinnunen, Matti Nelimarkka \& Reeta Pöyhtäri. 2020.
\newblock The {Datafication} of {Hate}: {Expectations} and {Challenges} in {Automated} {Hate} {Speech} {Monitoring}.
\newblock \emph{Frontiers in Big Data} 3.
\newblock \doi{10.3389/fdata.2020.00003}.

\bibitem[{Lee et~al.(2023)Lee, Jung \& Oh}]{lee_hate_2023}
Lee, Nayeon, Chani Jung \& Alice Oh. 2023.
\newblock Hate {Speech} {Classifiers} are {Culturally} {Insensitive}.
\newblock In Sunipa Dev, Vinodkumar Prabhakaran, David Adelani, Dirk Hovy \& Luciana Benotti (eds.), \emph{Proceedings of the {First} {Workshop} on {Cross}-{Cultural} {Considerations} in {NLP} ({C3NLP})}, 35--46. Dubrovnik, Croatia: Association for Computational Linguistics.
\newblock \doi{10.18653/v1/2023.c3nlp-1.5}.

\bibitem[{Liu et~al.(2019)Liu, Ott, Goyal, Du, Joshi, Chen, Levy, Lewis, Zettlemoyer \& Stoyanov}]{liu_roberta_2019}
Liu, Yinhan, Myle Ott, Naman Goyal, Jingfei Du, Mandar Joshi, Danqi Chen, Omer Levy, Mike Lewis, Luke Zettlemoyer \& Veselin Stoyanov. 2019.
\newblock {RoBERTa}: {A} {Robustly} {Optimized} {BERT} {Pretraining} {Approach}.
\newblock \doi{10.48550/arXiv.1907.11692}.

\bibitem[{van~der Nagel(2018)}]{van_der_nagel_networks_2018}
van~der Nagel, Emily. 2018.
\newblock ‘{Networks} that work too well’: intervening in algorithmic connections.
\newblock \emph{Media International Australia} 168(1). 81--92.
\newblock \doi{10.1177/1329878X18783002}.

\bibitem[{Parker \& Ruths(2023)}]{parker_is_2023}
Parker, Sara \& Derek Ruths. 2023.
\newblock Is hate speech detection the solution the world wants?
\newblock \emph{Proceedings of the National Academy of Sciences} 120(10). e2209384120.
\newblock \doi{10.1073/pnas.2209384120}.

\bibitem[{Rawat et~al.(2024)Rawat, Kumar \& Samant}]{rawat_hate_2024}
Rawat, Anchal, Santosh Kumar \& Surender~Singh Samant. 2024.
\newblock Hate speech detection in social media: {Techniques}, recent trends, and future challenges.
\newblock \emph{WIREs Computational Statistics} 16(2). e1648.
\newblock \doi{10.1002/wics.1648}.

\bibitem[{{Royal Commission of Inquiry into the terrorist attack on Christchurch masjidain on 15 March 2019}(2020)}]{royal_commission_of_inquiry__into_the_terrorist_attack_on_christchurch_masjidain_on_15_march_2019_ko_2020}
{Royal Commission of Inquiry into the terrorist attack on Christchurch masjidain on 15 March 2019}. 2020.
\newblock Ko tā tātou kāinga tēnei: {Report} of the {Royal} {Commission} of {Inquiry} into the terrorist attack on {Christchurch} masjidain on 15 {March} 2019.
\newblock Tech. rep. Royal Commission of Inquiry into the terrorist attack on Christchurch masjidain on 15 March 2019 Wellington, New Zealand.

\bibitem[{Sap et~al.(2019)Sap, Card, Gabriel, Choi \& Smith}]{sap_risk_2019}
Sap, Maarten, Dallas Card, Saadia Gabriel, Yejin Choi \& Noah~A. Smith. 2019.
\newblock The {Risk} of {Racial} {Bias} in {Hate} {Speech} {Detection}.
\newblock In Anna Korhonen, David Traum \& Lluís Màrquez (eds.), \emph{Proceedings of the 57th {Annual} {Meeting} of the {Association} for {Computational} {Linguistics}}, 1668--1678. Florence, Italy: Association for Computational Linguistics.
\newblock \doi{10.18653/v1/P19-1163}.

\bibitem[{Steen et~al.(2023)Steen, Yurechko \& Klug}]{steen_you_2023}
Steen, Ella, Kathryn Yurechko \& Daniel Klug. 2023.
\newblock You {Can} ({Not}) {Say} {What} {You} {Want}: {Using} {Algospeak} to {Contest} and {Evade} {Algorithmic} {Content} {Moderation} on {TikTok}.
\newblock \emph{Social Media + Society} 9(3).
\newblock \doi{10.1177/20563051231194586}.

\bibitem[{Tirrell et~al.(2025)Tirrell, Beaver, Stanley, Alqassas, Grieser, Conner \& Baptista}]{tirrell_language_2025}
Tirrell, Lynne, David Beaver, Jason Stanley, Ahmed Alqassas, Jessi Grieser, Tracy Conner \& Marlyse Baptista. 2025.
\newblock Language, conflict, and peace-making: contributions from linguistics and the philosophy of language.

\bibitem[{Tontodimamma et~al.(2021)Tontodimamma, Nissi, Sarra \& Fontanella}]{tontodimamma_thirty_2021}
Tontodimamma, Alice, Eugenia Nissi, Annalina Sarra \& Lara Fontanella. 2021.
\newblock Thirty years of research into hate speech: topics of interest and their evolution.
\newblock \emph{Scientometrics} 126(1). 157--179.
\newblock \doi{10.1007/s11192-020-03737-6}.

\bibitem[{Vidgen \& Derczynski(2020)}]{vidgen_directions_2020}
Vidgen, Bertie \& Leon Derczynski. 2020.
\newblock Directions in abusive language training data, a systematic review: {Garbage} in, garbage out.
\newblock \emph{PLOS ONE} 15(12). e0243300.
\newblock \doi{10.1371/journal.pone.0243300}.

\bibitem[{Waseem(2016)}]{waseem_are_2016}
Waseem, Zeerak. 2016.
\newblock Are {You} a {Racist} or {Am} {I} {Seeing} {Things}? {Annotator} {Influence} on {Hate} {Speech} {Detection} on {Twitter}.
\newblock In David Bamman, A.~Seza Doğruöz, Jacob Eisenstein, Dirk Hovy, David Jurgens, Brendan O'Connor, Alice Oh, Oren Tsur \& Svitlana Volkova (eds.), \emph{Proceedings of the {First} {Workshop} on {NLP} and {Computational} {Social} {Science}}, 138--142. Austin, TX: Association for Computational Linguistics.
\newblock \doi{10.18653/v1/W16-5618}.

\bibitem[{Wong(2024{\natexlab{a}})}]{wong_sociocultural_2024}
Wong, Sidney. 2024{\natexlab{a}}.
\newblock Sociocultural {Considerations} in {Monitoring} {Anti}-{LGBTQ}+ {Content} on {Social} {Media}.
\newblock In Vinodkumar Prabhakaran, Sunipa Dev, Luciana Benotti, Daniel Hershcovich, Laura Cabello, Yong Cao, Ife Adebara \& Li~Zhou (eds.), \emph{Proceedings of the 2nd {Workshop} on {Cross}-{Cultural} {Considerations} in {NLP}}, 84--97. Bangkok, Thailand: Association for Computational Linguistics.
\newblock \doi{10.18653/v1/2024.c3nlp-1.7}.

\bibitem[{Wong(2024{\natexlab{b}})}]{wong_what_2024}
Wong, Sidney Gig-Jan. 2024{\natexlab{b}}.
\newblock What is the social benefit of hate speech detection research? {A} {Systematic} {Review}.
\newblock In Daryna Dementieva, Oana Ignat, Zhijing Jin, Rada Mihalcea, Giorgio Piatti, Joel Tetreault, Steven Wilson \& Jieyu Zhao (eds.), \emph{Proceedings of the {Third} {Workshop} on {NLP} for {Positive} {Impact}}, 1--12. Miami, Florida, USA: Association for Computational Linguistics.
\newblock \doi{10.18653/v1/2024.nlp4pi-1.1}.

\bibitem[{Xiang et~al.(2012)Xiang, Fan, Wang, Hong \& Rose}]{xiang_detecting_2012}
Xiang, Guang, Bin Fan, Ling Wang, Jason Hong \& Carolyn Rose. 2012.
\newblock Detecting offensive tweets via topical feature discovery over a large scale twitter corpus.
\newblock In \emph{Proceedings of the 21st {ACM} international conference on {Information} and knowledge management} {CIKM} '12, 1980--1984. New York, NY, USA: Association for Computing Machinery.
\newblock \doi{10.1145/2396761.2398556}.

\bibitem[{Zhou et~al.(2023)Zhou, Cabello, Cao \& Hershcovich}]{zhou_cross-cultural_2023}
Zhou, Li, Laura Cabello, Yong Cao \& Daniel Hershcovich. 2023.
\newblock Cross-{Cultural} {Transfer} {Learning} for {Chinese} {Offensive} {Language} {Detection}.
\newblock In Sunipa Dev, Vinodkumar Prabhakaran, David Adelani, Dirk Hovy \& Luciana Benotti (eds.), \emph{Proceedings of the {First} {Workshop} on {Cross}-{Cultural} {Considerations} in {NLP} ({C3NLP})}, 8--15. Dubrovnik, Croatia: Association for Computational Linguistics.
\newblock \doi{10.18653/v1/2023.c3nlp-1.2}.

\end{thebibliography}
\end{document}